\documentclass[preprint]{elsarticle}

\usepackage{hyperref,}
\usepackage{bm}
\usepackage{booktabs}
\usepackage{multirow}
\usepackage{collcell}

\journal{Journal of \LaTeX\ Templates}

\usepackage{amssymb}
\usepackage{amsmath}
\usepackage{tikz}

\begin{document}
\usetikzlibrary{math}

\begin{frontmatter}

\title{Combining Graph Neural Networks and Spatio-temporal Disease Models to Predict COVID-19 Cases in Germany}

\author[lmu]{Cornelius Fritz\corref{sha}}
\author[lmu,icb]{Emilio Dorigatti\corref{sha}}
\author[lmu]{David R{\"u}gamer\corref{corr}}

\address[lmu]{Department of Statistics, Ludwig Maximilian Universit{\"a}t, M{\"u}nchen, Germany}
\address[icb]{
Helmholtz Zentrum M{\"u}nchen – German Research Center for Environmental Health, Neuherberg, Germany}


\cortext[sha]{Equal contribution}
\cortext[corr]{Corresponding author}



\begin{abstract}
During 2020, the infection rate of COVID-19 has been investigated by many scholars from different research fields. In this context, reliable and interpretable forecasts of disease incidents are a vital tool for policymakers to manage healthcare resources. Several experts have called for the necessity to account for human mobility to explain the spread of COVID-19. Existing approaches are often applying standard models of the respective research field. This habit, however, often comes along with certain restrictions. For instance, most statistical or epidemiological models cannot directly incorporate unstructured data sources, including relational data that may encode human mobility. In contrast, machine learning approaches may yield better predictions by exploiting these data structures, yet lack intuitive interpretability as they are often categorized as black-box models. We propose a trade-off between both research directions and present a multimodal learning approach that combines the advantages of statistical regression and machine learning models for predicting local COVID-19 cases in Germany. This novel approach enables the use of a richer collection of data types, including mobility flows and colocation probabilities, and yields the lowest MSE scores throughout our observational period in our benchmark study. The results corroborate the necessity of including mobility data and showcase the flexibility and interpretability of our approach.
\end{abstract}


\end{frontmatter}


\section{Introduction}

In December 2019, the region of Wuhan, China, experienced an outbreak of a novel coronavirus, initially infecting around 40 people \citep{Wu2020}. Impeded by the early findings that patients are already infectious in the pre-symptomatic stage of the disease \citep{Rothe2020} and that transmission occurs mostly either through the exchange of virus-containing droplets \citep{Guan2020} or expiratory particles \citep{Asadi2020}, i.e., aerosols, the containment strategies were not sufficient in detaining the virus from becoming a pandemic. Consequentially, the World Health Organization declared the virus a pandemic in March 2020 with 66.2 million infections and 1.5 million deaths worldwide as of December 7th, 2020 \citep{Organization2020}.


Given this development, it was repeatedly pondered how and if mathematical modeling can help contain the current COVID-19 crisis \citep{Panovska-Griffiths2020}. We argue that data science and machine learning can provide urgently needed tools to doctors and policymakers in various applications. For instance, model-assisted identification and localization of COVID-19 in chest X-rays can support doctors at achieving correct and precise diagnoses \citep{Wang2021}. Meanwhile, policymakers benefit from studies determining and evaluating specific policies  \citep{Kraemer2020,Chang2020a}. In one noteworthy example, \citep{Flaxman2020} were able to quantify to what extend targeted non-pharmaceutical interventions (NPIs) aided in ceasing the exponential growth  rate of COVID-19. This work is often quoted as the main driver of the social distancing measures implemented in the UK, thus allowing the British government to pursue evidence-based containment strategies. 

In order to adequately evaluate the role of specific policies and implement a successful containment strategy, a robust and interpretable forecast of the pandemic's state into the future is necessary. Among other purposes, this endeavor allows authorities to better manage healthcare resources (hospital beds, respirators, vaccines etc.). 

The corresponding modeling task is broad and can be tackled at various levels of spatio-temporal granularity. While some proposals operate on daily country-level data \citep{Zeroual2020,Stubinger2020}, others are designed to provide local predictions \citep{Liu2020,DeNicola2020}. From a methodological point of view, most of these approaches are influenced by either epidemiology, time-series analysis, regression models, machine or deep learning. However, we argue that the most promising approaches combine ideas from seemingly distant research areas with new types of data. Thereby, one can bypass restrictions of simpler models and improve the model's forecasting performance while also benefiting from the merits of each respective idea. In addition, allowing for the inclusion of novel data modalities in some of the more traditional approaches may further improve models. This was highlighted in previous literature by \citep{Oliver2020, Buckee2020a}, who suggest to include non-standard data sources, e.g., aggregated contact patterns obtained from mobile phones or behavioral data, into the analysis to help in understanding and fighting COVID-19. 

\paragraph{Hybrid modeling approaches} Examples of these types of hybrid models are scattered throughout the literature. \citep{Liu2020} combine a mechanistic metapopulation commonly used in epidemiology with clustering and data augmentation techniques from machine learning to improve their forecasting performance. For this endeavor, additional data sources, including news reports and internet search activity, were leveraged to inform the global epidemic and mobility model, an epidemiological model already successfully applied to the spread of the Zika virus \citep{Zhang2017a}. In another notable application, \cite{Chang2020a} enrich a relatively simple metapopulation model with mobility flows to numerous points of interest. Subsequently, this model is used to predict the effect of reopening after a specific type of lock-down through counterfactual analysis. With the help of Facebook, \cite{Fritz2020} utilize mobility and aggregated friendship networks to discover how these networks drive the infection rates on the local level of federal districts in Germany. 
In this context, another route to accommodate for such network data is employing a graph neural network (GNN). This technique builds on the intuitive idea of message passing \cite{Gilmer2017} and has recently attracted a lot of attention in the deep learning community. For graph neural networks there is a wide range of possible applications, namely node classification or forecasting \citep{Wu2019,Bronstein2017}. In several occasions, this model class has already been applied to forecast the number of COVID-19 cases in 2020.

\paragraph{Applying GNNs to COVID-19 data} \cite{panagopoulosTransferGraphNeural2020} construct a graph whose edges encode mobility data for a given time point collected from Facebook. Their approach exploits a long short-term memory (LSTM) architecture to aggregate latent district features obtained from several graph convolutional layers and transfer learning that accounts for the asynchrony of outbreaks across borders.
In a comparable proposal, \cite{gaoSTANSpatioTemporalAttention2020} employ a graph neural network to encode spatial neighborhoods and a recurrent neural network to aggregate information in the temporal domain.
Through a novel loss function, they simultaneously penalize the squared error of the predicted infected and recovered cases as well as the long-term error governed by the transmission and recovery rates within traditional Susceptible-Infectious-
Recovered (SIR) models. Contrasting these approaches, 
\cite{dengColaGNNCrosslocationAttention2020} propose a recurrent neuronal network to derive latent features for each location, hence they construct a graph whose edge weights are given by a self-attention layer.
Instead of using a recurrent neural network, \cite{kapoorExaminingCOVID19Forecasting2020} construct a spatio-temporal graph by creating an augmented spatial graph that includes all observed instances of the observed network side-by-side and enables temporal dependencies by connecting each location with the corresponding node in previous days.

\paragraph{Our contribution}

In this work, we propose a novel fusion approach that directly combines dyadic mobility and connectedness data derived from the online platform Facebook with structural and spatial information of Germany's cities and districts. In contrast to \cite{Fritz2020}, we learn each district's embedding in the network in an end-to-end fashion, thus there is no need for a separate pre-processing step. 
With this, we heed recent calls such as \cite{Oliver2020, Buckee2020a} highlighting the need for more flexible and hybrid approaches taking also dyadic sources of information into account.
From a methodological point of view, we make this possible by combining graph neural networks with epidemiological models \cite{Held2012, Held2017} to simultaneously account for network-valued data and tabular data. We further provide comparisons, sanity checks as well as uncertainty quantification to investigate the reliability of our proposed model. In our application case, we provide forecasts of weekly COVID-19 cases with disease onset at the local level of 401 federal districts in Germany as provided by the Robert-Koch Institute.\\

\noindent This article's remainder is structured as follows. In the following section, the employed data are described in detail to determine several caveats that need to be accounted for when working with official surveillance data of an ongoing pandemic. Consecutively, Section~\ref{sec:related} lays out the groundwork of distributional regression and graph neural networks that we combine in our main contribution, proposed in Section~\ref{sec:main}. To ascertain the practical use of this proposed model, Section~\ref{sec:eval} applies the proposed approach to Germany's weekly data. Finally, our article ends with a discussion and concluding remarks in Section~\ref{sec:discussion}.     


\section{Data}
\label{sec:data}

We distinguish our data sources between Facebook and infection data. While the infection data are time-series that are solely utilized in our model's structured and target component, most of the network data are directly used in the GNN module. To allow for sanity checks and interpretable coefficients, the networks are also transformed onto the units where we measured the time-series by calculating specific structural characteristics from the networks following \citep{Fritz2020}.

\subsection{Facebook data on human mobility and connectedness}

To quantify the social and mobility patterns on the regional level, we use data on friendship ties, colocation probabilities, and district-specific data on the proportion of people staying put provided by Facebook \citep{Maas}. These spatial data sets were made available through the \textsl{Data for Good} program and used in several other publications \citep{Bonaccorsi2020,Fritz2020, Lorch2020,Holtz,Galeazzi2020}. 
The spatial units are on the NUTS 3 level and encompass $n = 401$ federal districts. All data were collected from individual mobile phone location traces of Facebook users that opted in the \emph{Location History} setting on the mobile Facebook application. For data security reasons, these individual traces were subject to differential privacy and aggregated to the 401 federal districts. As a result, the nodes of all networks are these districts. We augment the given network data with spatial networks encoding neighboring districts and distances in kilometers, that are respectively denoted by $x^N \in \lbrace 0,1 \rbrace^{n\times n}$ and $x^D \in \mathbb{R}_{>0}^{n\times n}$.

\paragraph{Colocation networks}
The first type of network data that we incorporate in our forecast are colocation maps, which indicate the probability of two random persons from two districts to meet one another during a given week. We obtain for each week $t$ a colocation matrix $x^{C}(t)\in \mathbb{R}_{>0}^{n \times n}$, where then $ij$th entry ($x_{ij}^{C}(t)$) indicates the probability of an arbitrary person from district $i$ to meet another person from district $j$. To incorporate such network-valued data in the structured part of our framework, we transform it to tabular data. More specifically, we follow \citep{Fritz2020} and use the Gini index to measure the concentration of meeting patterns of districts. For district $i \in \lbrace 1, ..., n\rbrace$ this index can be defined through: 
\begin{align*}
    x_{i}^G(t) = \frac{\sum_{m, n \neq i} |x_{im}^C(t) -x_{in}^C(t) |}{2 n \sum_{j\neq i} x_{ij}^C(t) } \in [0,1].
    \label{eq:gini}
\end{align*}
Higher values translate to a restricted meeting pattern within a district, while lower values suggest rather diffused social behavior. In our application, we perform a weekly standardization of $x_{i}^G(t)$. 

\paragraph{Social connectedness network}
Secondly, we quantify social connections between the districts via Facebook friendships. Utilizing information from a snapshot of all Facebook connections within Germany of April 2020, we derive a Social Connectedness Index, that was first introduced by \cite{Bailey2018a}. This pairwise time-constant index relates to the relative friendship strength between two districts and is stored in a weighted network $x^S \in \mathbb{R}_{>0}^{n \times n}$. For $i,j \in \lbrace 1, \ldots, n \rbrace$ the entries of $x^S$ are given by: 
	\begin{align*}
	x_{ij}^S= \frac{\# \lbrace \text{Friendship Ties between users in district $i$ and $j$} \rbrace}{\# \lbrace \text{Users in district $i$} \rbrace \# \lbrace \text{Users in district $j$} \rbrace}.
	\end{align*}
Via multidimensional scaling, we can obtain two-dimensional district-specific embeddings from $x^S$, which we denote by $x_i^S$ and incorporate in the structured component.

\paragraph{Staying put}
Besides, we incorporate the weekly percentage of people staying put as a measure for people's compliance (Facebook users) with social distancing. We derive the corresponding weekly district-specific measure $x_i^{SP}(t)$ by averaging daily measures provided by Facebook. In this context, $\textsl{staying put}$ is defined as being observed in one $0.6km \times 0.6km$ square throughout a day \cite{Facebook}. 
 
\subsection{Time-series of daily COVID-19 infections}

Current data on the pandemic's state in Germany are provided by the \href{https://hub.arcgis.com/datasets/dd4580c810204019a7b8eb3e0b329dd6_0/}{Robert-Koch-Institute}. From this database, we obtain the number of people with symptom onset and registered cases of COVID-19 grouped by age, gender, and federal district (NUTS-3 level) for each day. Due to the observation that mainly people aged between 15 and 59 years are active users of Facebook, we limit our analysis to the corresponding age groups, namely, people with age between 15-35 and 36-59. 

As discussed in \cite{Gunther2020}, the central indicator of the infection occurrence is the number of people with disease onset at a specific day; hence our application focuses on the respective quantity. Due to mild courses of infection and inconsistent data collection, the data of disease onset is not known in about 30$\%$ of the cases. Therefore, we impute the missing values by learning a probabilistic model of the delay between disease onset and registration date. Eventually, we attain $y_{ig}(t)$, the infection counts of district $i$ and group $g$ during week $t$, by sampling from the estimated distribution of delay times. By doing that, we follow \cite{Fritz2020} where the procedure is given in more detail. The groups $g$ are elements of the Cartesian product of all possible age and gender groups. We denote all corresponding features by the vector $\bm{x}_{ig}(t)$.

Further, we are given data on the population sizes of each group $g$ and district $i$ from the German Federal Statistical Office, denoted by $pop_{ig}$, based on which we compute the population density $den_{ig}$. In this setting, one can assume that not the count but rate of infections in a specific district and group carries vital information. Hence, the target we model is the corresponding rate defined by  $\tilde{y}_{ig}(t) = \frac{y_{ig}(t)}{pop_{ig}}$. 

\section{Methodological background}
\label{sec:related}

First,  we present the methodological foundations of our approach: distributional regression and graph neural networks. To begin with, we will introduce distributional regression approaches on the basis of which we define our deep probabilistic model in Section~\ref{sec:main}.  To incorporate modeling techniques from statistics and epidemiology in our network, we use so-called structured additive predictors that represent smooth additive effects of input features and can be represented in a neural network. As we will elaborate in our main section, effect smoothness can be achieved by a specifically designed network regularization term. The second building block of our model are graph neural networks, which we introduce subsequently.  

\subsection{Distributional regression} \label{sec:dr}

Distributional regression is a modeling approach to regress a (parametric) distribution $\mathcal{D}$ on $p$ given input features $\bm{x} \in \mathbb{R}^p$ \cite{Koenker.2013}. In contrast to other regression approaches that, e.g., do only relate the mean of an outcome variable to certain features, distributional regression also accounts for the uncertainty of the data distribution, known as aleatoric uncertainty \citep{Horal.1996}. 
Given a parametric distributional assumption $\mathcal{D}(\theta_1,\ldots,\theta_K)$, the model learns the distributional parameters $\bm{\theta} = (\theta_1,\ldots,\theta_K)^\top$ by means of feature effects. In structured additive distributional regression \cite{Klein.2015}, each distributional parameter is estimated using an additive predictor $\eta_k(\bm{x}_k)$. In this context, $\eta_k: \mathbb{R}^{p_k} \to \mathbb{R}$ is an additive transformation of a pre-specified set of features $\bm{x}_k \in \mathbb{R}^{p_k}$, $1<p_k<p$. This additive predictor is finally transformed to match the domain of the respective parameter by a monotonic and differentiable transformation function $h_k$: $$\theta_k(\bm{x}_k) = h_k\big(\eta_k(\bm{x}_k)\big).$$ Note that the $K$ parameters relating to $\mathcal{D}$ now depend on the features. 

Moreover, structured additive distributional regression models 
allow for a great variety of feature effects in the additive predictor that are interpretable by nature \cite{Wood.2017}. Examples include linear, non-linear, or random effects of one or more features. The latter two effect types can be represented via basis functions (such as regression splines, polynomial bases or B-splines). Further examples and details can be found, e.g., in \cite{Ruppert.2003, Wood.2017}. 
Due to the additivity of effects in $\eta_k$, the influence of single features or a combination of features can in many cases be directly related to the mean or the variance of the modeled distribution. 



\paragraph{Semi-structured deep distributional regression}
\label{sec:SDDR}
A recent trend is the combination of neural networks with statistical regression models in various ways \citep{De.2011, Cheng.2016, Bras.2019, Agarwal.2020, Poelsterl.2020, Ruegamer.2020b, Baumann.2020}. In this work, we make use of \textit{semi-structured deep distributional regression} \citep[SDDR,][]{Ruegamer.2020}. SDDR combines neural networks and structured additive distributional regression by embedding the statistical regression model in the neural network and ensures the identifiability of the regression model part. SDDR is a natural extension of distributional regression by extending the additive predictor $\eta_k$ of each distributional parameter with its \emph{structured effects} to latent features, so-called \emph{unstructured effects}, that are learned by one or more deep neural network (DNN). To disentangle the structured model parts from the unstructured parts, an orthogonalization cell is used to ensure identifiability of the structured model effects. 

\subsection{Graph neural networks}

Graphs are a mathematical description of the entities and their relationships in a given domain and naturally arise in a variety of seemingly distant fields.
However, due to their non-euclidean structure, it is not straightforward to apply traditional machine learning methods to problems involving graphs, as these methods operate on vectors \cite{Bronstein2017}.
A number of methods have been proposed to embed graphs into low-dimensional euclidean spaces, allowing to use the resulting vector representations for prediction tasks with traditional machine learning algorithms such as node classification and missing link prediction \cite{Wu2019}.
Inspired by the success of convolutional neural networks, comparable approaches formulated convolutions for graphs via the spectral domain relying on the convolution theorem \cite{bruna2013spectral}. Later versions of convolutional operators were adapted to the vertex domain by means of a message-passing framework \citep{Gilmer2017}. In this framework the feature vector of each node is updated based on the feature vectors of its own neighbors and the edges connecting them \cite{Zhang2019}. Following this work, more advanced neighborhood aggregation methods \cite{graphattn}, scalable inference \cite{graphsage} and domain-specific applications \cite{Gilmer2017} have been proposed.
In general, a graph neural network performs $R$ rounds of message passing, after which all nodes' latent features can be combined to obtain a unified representation for the whole graph \cite{mlg2017_21}, or individual nodes \cite{Kipf2017}. We can use the latter type of representations to derive node-specific predictions. 

Following the notation in \cite{Gilmer2017}, in each message-passing round $r$ the neighbors $N(v)$ of $v$ are aggregated into a message $\bm{m}_v^r$ through a message function $M_r$:
\begin{equation}\label{eq:gnn_message}
\bm{m}_v^r=\sum_{w\in N(v)} M_r(\bm{x}_v^r, \bm{x}_w^r, \bm{e}_{vw})
\end{equation}
where $\bm{x}^r_v$ denotes the latent features of node $v$ after $r$ rounds of message passing and $\bm{e}_{vw}$ the features associated to the edge connecting $v$ and $w$.
Next, an update function $U_r$ combines $\bm{x}_v^r$ and $\bm{m}_v^r$ to obtain the updated latent features $\bm{x}_v^{r+1}$:
\begin{equation}\label{eq:gnn_update}
\bm{x}_v^{r+1}=U_r(\bm{x}_v^r, \bm{m}_v^r)
\end{equation}
Taken together \eqref{eq:gnn_message} and \eqref{eq:gnn_update} define a message passing round that propagates information one hop further than the previous round. Multiple message passing rounds can be applied successively to diffuse information across the complete network.

\section{Combining network-valued and spatio-temporal disease data}
\label{sec:main}

\begin{figure*}[!t]
    \centering
    \vspace{0.3cm}
    \includegraphics[page=3,
    trim=1cm 1cm 0cm 1cm,, 
    width = \textwidth]{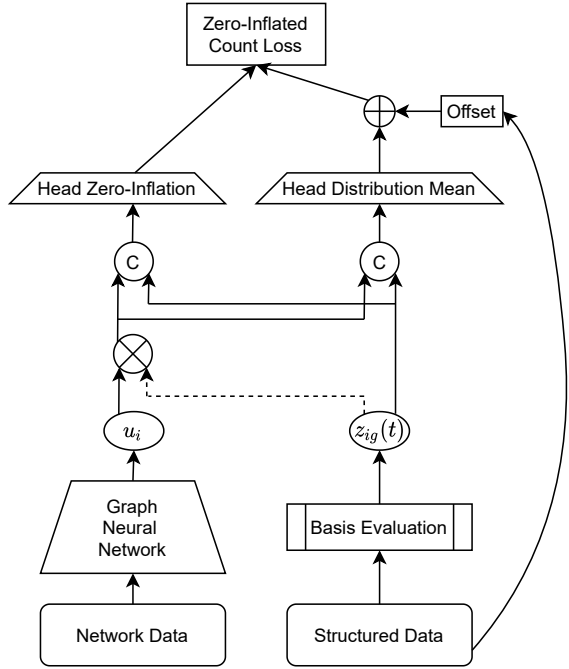}
    \caption{Network architecture of our proposed model. The mobility data is fed into a GNN on the bottom left to learn latent features $\bm{u}_i$. On the bottom right side, the structured data is transformed using basis evaluation. Using the orthogonalization the learned effects $\bm{u}_i$ are projected onto the orthogonal complement of selected parts of the basis evaluated structured features. Both parts are combined using a concatenation and fed into both a network head that learns the zero-inflation as well as a network head to learn the distribution's mean. After adding an offset to the mean, both parts are finally combined in a distributional layer that learns the zero-inflated count distribution based on the corresponding log-likelihood.}
    \label{fig:arch}
\end{figure*}

We present our hybrid modeling approach to forecast weekly district-wise COVID-19 cases based on structured and unstructured data sources described in Section~\ref{sec:data}. 
The general framework is depicted in Figure~\ref{fig:arch} and fuses the interpretability of distributional regression with a GNN architecture to flexibly learn all district's latent representation from the network data. 


\subsection{Neural network formulation} \label{sec:nn_form}

\paragraph{Distributional assumption}
Our considered time window stretches over a low-infection phase
during which 20 - 30\% of the observations reported no cases. This phase is delimited by two primary infection waves, approximately from March - April and October - November, showing more than 1,200 cases per week. Hence, our goal is to build a model that can adequately predict high numbers as well as zero observations, which are common during low-infection seasons. We achieve this by assuming that the cases follow a mixture distribution of a point mass distribution at zero and an arbitrary count distribution with mixture weight $\pi \in [0,1]$. Here, we regard any probability distribution defined over non-negative integers as a count distribution, e.g., the negative binomial, Poisson, and generalized Hermite distribution \cite{Puig2006}. 
The resulting \emph{zero-inflated} distribution $\mathcal{D}$ is primarily characterized by the mean of the count component $\lambda$ and the zero-inflation probability $\pi$. Any additional parameters relating to other traits of the count distribution, e.g., the scale parameter of the negative binomial distribution, are denoted by $\chi$. The probability mass function of $\mathcal{D}$ is given by: 
\begin{equation} \label{eq:zi_pmf}
    f_{\mathcal{D}}(y |\lambda, \pi, \chi) = \pi I(y = 0) + (1-\pi) f_{\mathcal{C}}(y|\lambda,\chi), 
\end{equation}
where $f_{\mathcal{C}}$ is the density of the chosen count distribution $\mathcal{C}$ and $I$ denotes the indicator function. By incorporating the point mass distribution, the model can capture excess rates of zero observations and $\pi$ may be interpreted as the percentage of excess-zero observation \cite{Lambert1992}. 

In our modeling approach for COVID-19 cases $y_{ig}(t)$ we relate structured data as well as network data to the parameters $\lambda$ and $\pi$ of the zero-inflated distribution, which yields the following distributional and structural assumption:
\begin{equation} \label{eq:modeldef}
\begin{gathered}
    y_{ig}(t) \sim \mathcal{D}(\lambda_{ig}(t),\pi_{ig}(t), \chi),\\ \lambda_{ig}(t) = h_1\left(\eta_{1,ig}(t)\right),\,\, \pi_{ig}(t) = h_2(\eta_{2,ig}(t)),
\end{gathered}
\end{equation}
with chosen zero-inflated distribution $\mathcal{D}$. For the structural component in \eqref{eq:modeldef}, the two feature-dependent distributional parameters are described through the additive predictors $\eta_{1,ig}(t)$ and $\eta_{2,ig}(t)$. We transform these predictors via fixed transformation functions $h_1,h_2$ to guarantee correct domains of the respective modeled parameter, e.g., a sigmoid function for the probability $\pi_{ig}(t)$ (see Section~\ref{sec:dr}). 

\paragraph{Additive predictors}

Inspired by SDDR \cite{Ruegamer.2020}, we estimate additive effects of tabular features on the parameters characterizing the zero-inflated distribution using a single-neuron hidden layer. As proposed in various statistical COVID-19 modeling approaches \cite{Fritz2020,Schneble2020a}, we learn these structured effects with an appropriate regularization to enforce smoothness of non-linear effects. This penalization can be seen as a trade-off between complexity and interpretability \cite{Wood2020}.

The additive predictors $\eta_{1,ig}(t)$ and $\eta_{2,ig}(t)$ can be defined in terms of both unstructured and structured features (left and right bottom input in Figure~\ref{fig:arch}). In the structured model part, we use the complete suite of district-specific features, defined in Section~\ref{sec:data}, including them as arbitrary additive effects, which are detailed in the next section. In the following, we make this clear by using $\bm{z}_{ig}(t)$, which are the input features $\bm{x}_{ig}(t)$ but transformed using some basis function evaluation. We denote the corresponding feature weights by $\bm{\vartheta}^{\mbox{\footnotesize{str}}} = ({\bm{\vartheta}_1^{\mbox{\footnotesize{str}}}}^\top,{\bm{\vartheta}_2^{\mbox{\footnotesize{str}}}}^\top)^\top$ corresponding to $\eta_1$ and $\eta_2$, respectively.
The unstructured part of our network computes each district's embedding (node) by exploiting time-constant district population attributes and edge attributes. For our application, these attributes encode geographic connectedness between districts and social connectedness. 
Here, the message passing framework enables the embeddings to contain first, second, and higher-order information about the spread of the disease among all districts.
Finally, we inform either additive predictors with the embeddings $\bm{u}_i$ for district $i$ learned from the GNN to incorporate the network data in the distributional framework. 

\paragraph{Orthogonalization} Identifiability is crucial to our analysis since some feature information is shared between the structured effects and unstructured effects. For instance, the social connectedness index $x^S$ is exploited in the structured part via the MDS-embeddings but also in the graph neural network as an edge attribute. If these two model parts are not adequately disentangled, it is unclear what part of the model is accounting for which information in the shared features. 
Therefore, the latent GNN representations $\bm{u}_i$ are  orthogonalized with respect to $\bm{z}_{ig}(t)$ yielding $\tilde{\bm{u}}_i = \bm{u}_i \otimes \bm{z}_{ig}(t)$. More specifically, we project $\bm{u}_i$ in the orthogonal complement of the column space spanned by $\bm{z}_{ig}(t)$ and use $\tilde{\bm{u}}_i$ instead of $\bm{u}_i$ in the additive predictors (see \cite{Ruegamer.2020} for further details). 
In the final step, we combine the structured and unstructured effects as a sum of linear orthogonalized embedding effects and the structured predictors, i.e., $\eta_{k,ig} = \bm{x}_{ig}(t) \bm{\vartheta}_k^{\mbox{\footnotesize{str}}} + \tilde{\bm{u}}_k \bm{\vartheta}_k^{\mbox{\footnotesize{unstr}}}$, $k \in \{1,2\}$.

\paragraph{Different exposures in count regression}
As already discussed in Section~\ref{sec:data}, the primary goal of our application is modeling the rates of infections rather than the raw observed counts, as each state is subject to a different exposure of COVID-19. Still we learn the mean of the counting distribution in \eqref{eq:modeldef} and correct for the differing population sizes by adding a constant offset term to the concatenated linear predictor, i.e., we subtract $\log(pop_{ig})$ of $\eta_{1,ig}(t)$. For additional information on this procedure in the realm of zero inflated models, we refer to \cite{Lee2001}.




\subsection{Proposed COVID-19 model specification} \label{sec:model_covid}

\begin{table}[t!]
    \centering
    \begin{tabular}{l|c | c c}
\multirow{2}{*}{Feature} & \multicolumn{1}{c|}{Structured Part} & \multicolumn{2}{c}{GNN} \\ 
  &   Basis Evaluation (Transformation) & Node & Edge \\\hline
          $\tilde{y}_{ig}(t-1)$  & TP-Spline ($\mbox{logp1}(\cdot)$) &  &   \\
           $x_i^G(t)$ &  Bivariate Spline with $t$ &  &   \\
           $x^S$ &  - &  & \checkmark \\
            $x_i^S$ &  Bivariate Spline &  &   \\
           $x^D$ & - &  & \checkmark   \\
           $x^N$ & - &  & \checkmark \\
           $x_i^{SP}(t)$   & Bivariate Spline with $t$ &  &   \\
           $pop_{ig}$ & Offset ($\log(\cdot)$) & \checkmark &   \\
           $den_{ig}$ & - & \checkmark & \\ 
           $g$ & Dummy Variables & & \\
    \end{tabular}
    \caption{Features and their incorporation into the structured and GNN part of our model. For each feature, the second column indicates the basis function evaluation used in the structured part, which is applied to the the feature itself or a transformation of it given in brackets. If no transformation is given, the identity is used. For bivariate effects we use bivariate thin plate (TP) regression splines. The transformation $\mbox{logp1}(y) = \log(y + 1)$. The third column indicates the incorporation of each feature in the GNN part, either as a node or edge feature. To also account for the group-specific nature of each distributional parameter, we further add a gender and age effect using $g$ as a dummy variable.}
    \label{tab:features_in_model}
\end{table}
\paragraph{Distributional regression}
On the basis of \eqref{eq:modeldef}, we set up our COVID-19 model as follows: let $\mathcal{D}$ define a zero-inflated Poisson (ZIP) distribution. We reparameterize the distribution in terms of only one additive predictor $\eta_{ig}(t) = \eta_{1,ig}(t)$ as this proved to help numerical stability. We therefore define $\pi_{ig}(t) \equiv \exp(-\exp(\chi + \log(\lambda_{ig}(t))))$ and learn the distribution's rate via $\lambda_{ig}(t) = \exp(\eta_{ig}(t))$. 

Another option for modeling counts is to use a negative binomial (NB) distribution as, e.g., done by \cite{Fritz2020}. The NB distribution is often chosen over the Poisson distribution due to its greater flexibility, particularly by allowing to account for overdispersion. We will compare the NB distribution against the ZIP approach by reparameterizing the NB distribution in terms of its mean, similarly to what did for the ZIP. Table~\ref{tab:features_in_model} further summarizes the use of all available features in our model using their transformation and incorporation in the structured additive as well as GNN model part.


\paragraph{Graph neural network}
Among all possible variants of the general message passing framework described above, we opt for the proposition of \cite{Monti2017}, thereby making use of edge attributes and efficiently handling our relatively large graphs.
The message function of \eqref{eq:gnn_message} thus becomes:
\begin{equation}
M_r(\bm{x}_v^r, \bm{x}_w^r, \bm{e}_{vw})
=\frac{1}{H^r\cdot|N(v)|}
\sum_{h=1}^{H^r} w_h^r(\bm{e}_{vw}) \cdot \bm{\Theta}_h^r \bm{x}_w^r
\end{equation}
which uses $H^r$ linear maps to transform the neighbors' features and $H^r$ radial basis function (RBF) kernels to weight the linear maps:
\begin{equation} \label{eq:gconv2}
w_h^r(\bm{e})=\exp\left\lbrace-\frac 1 2 (\bm{e}-\bm{\mu}_h^r)^\top \text{diag}({\bm{\sigma}_h^r}^2)^{-1}(\bm{e}-\bm{\mu}_h^r)\right\rbrace.
\end{equation}
Whereas the node update function of \eqref{eq:gnn_update} becomes:
\begin{equation}
U_r(\bm{x}_v^r, \bm{m}_v^r)=\bm{m}_v^r+\bm{\Theta}_0\bm{x}_v^r+\bm{b}^r
\end{equation}
with trainable parameters $\bm{\Theta}_0^r$, $\bm{\Theta}_1^r,\ldots,\bm{\Theta}_{H^r}^r$, $\bm{\mu}_h^r$, $\bm{\sigma}_h^r$ and $\bm{b}^r$.

Because people can travel from any district to any other district, e.g., via train or car, we use a fully connected graph as input to the GNN and embedded information about social connectedness in the edge attributes.
The first graph convolutional layer uses $H^1=8$ affine maps with output dimensionality of 256, followed by a the second layer that further reduces this number to 128 latent components with $H^2=4$.
Next, four fully-connected layers successively reduce the dimensionality of the node embeddings to 16 components.
All layers use batch normalization \citep{Ioffe2015} followed by leaky ReLU activation \citep{Maas13}. To reduce the chance of overfitting, we further use dropout \citep{Srivastava2014} with probability $0.25$ after the two graph convolutions.

\subsection{Uncertainty quantification} \label{sec:uq}

A crucial tool to investigate the model's reliability is to assess its uncertainty. While the proposed approach explicitly models the uncertainty in the given data distribution (aleatoric uncertainty), we can also derive epistemic uncertainty of parts of the model through its connection to statistical models.

\paragraph{Epistemic uncertainty}
Using standard regression model theory, we can derive the epistemic uncertainty of our model, i.e., the uncertainty of model's weights. When regarding the GNN part of the model as a fixed offset $\bm{o}$ and fixing the amount of smoothness defined by $\bm{\xi}$
, it follows 
\begin{equation}
\bm{\vartheta}^{\mbox{\footnotesize{str}}} \mid \bm{y}, \bm{\xi}, \bm{o} \sim \mathcal{N}(\hat{\bm{\vartheta}}, (\hat{\bm{\mathcal{I}}}^{\mbox{\footnotesize{str}}} + \bm{P})^{-1}),
\end{equation}
where $\hat{\bm{\mathcal{I}}}^{\mbox{\footnotesize{str}}}$ is the Hessian of the negative log-likelihood at the estimated network parameters $\hat{\bm{\vartheta}}$ \cite{Wood.2017} . 
We note that especially the conditioning on $\bm{o}$ (the GNN) neglects some additional variance in the parameter estimates but still allows us to get a feeling for the network's uncertainty.

To additionally account for the epistemic uncertainty of the GNN, we turn to deep ensembles \citep{lakshminarayanan2017simple}, a simple method known to result in reliable uncertainty estimates.
In this context, the epistemic uncertainty can be estimated simply by computing the standard error of the predictions of an ensemble of models, each trained from scratch with a different random initialization.

\paragraph{Aleatoric uncertainty} In addition to epistemic uncertainty, our model accounts for aleatoric uncertainty by modeling all the distributional parameters of the zero-inflated count distribution explicitly. For example, in the case of the ZIP distribution, the distribution's variance can be derived from its parameters as follows:
\begin{equation} \label{eq:aleatoric}
    (1 - \pi_{ig}(t)) \cdot (\lambda_{ig}^2(t) + \lambda_{ig}(t) \chi) - (1 - \pi_{ig}(t)) \cdot \lambda_{ig}(t)^2.
\end{equation}
In particular, this allows us to make a probabilistic forecast for all weeks, districts, and each cohort (age, gender). After having observed the forecasted values, we can assess how well the model performed and how well it predicted uncertainty of the data distribution when only trained with historic data up to a certain week.

\subsection{Network training}
We train model \eqref{eq:modeldef} by minimizing the negative log-likelihood derived from the distributional assumption in \eqref{eq:modeldef}. The combined $\rho := p_1 + p_2 + \tau + p_\chi$ weights $\bm{\vartheta} \in \mathbb{R}^{\rho}$ of the whole network subsume $p_1$ weights for the first additive predictor $\eta_1$, $p_2$ effects for the second additive predictor $\eta_2$, $\tau$ effects for the GNN and $p_\chi$ weights for additional distributional parameters. For readability, we omit in the following the indices $i$ and $g$ as well as the time-dependency of the two distributional parameters, yet make the dependency on learned weights explicit. Stemming from \eqref{eq:zi_pmf}, we can construct a joint likelihood  $\ell(\bm{\vartheta})$ by summing up the contribution of each individual observation, that is given by 
\begin{equation} \label{eq:likelihood}
    \ell(\bm{\vartheta}) = \log\left(\pi(\bm{\vartheta})  I(y = 0) + (1-\pi) f_{\mathcal{C}}(y|\lambda(\bm{\vartheta}),\chi) \right).
\end{equation}
Under conditional independence, feature weights can be learned by minimizing the sum of negative log-likelihood contributions for all observations. To avoid overfitting and estimate smooth additive effects in the structured part of our model, we add a quadratic penalty term $J(\bm{\vartheta}) = \sum_{j=1}^2  {\bm{\vartheta}_j^{\mbox{\footnotesize{str}}}}^\top \bm{P}_j \bm{\vartheta}_j^{\mbox{\footnotesize{str}}}$. Thereby, we regularize weights in the network $\bm{\vartheta}_j \in \mathbb{R}^{p_j}$ that correspond to smooth structured effects. 
Penalization is controlled by individual complexity parameters that we incorporate in the penalty matrices $\bm{P}_j \in \mathbb{R}^{p_j \times p_j}, j=1,\ldots,2$. These matrices, in turn, are block-diagonal matrices that are derived by the chosen basis functions in the structured model part \cite{Wood.2017}. 
We do not additionally penalize the count parameter $\chi$ nor the GNN model part other than using the orthogonalization. In practice, we observe that training the network can be stabilized when choosing RMSprop \citep{Tieleman.2012} as the optimizer.

\section{Application and evaluation}
\label{sec:eval}

We now apply our model from Section~\ref{sec:model_covid} to the data introduced in Section~\ref{sec:data}.
To this end, we first explain our evaluation scheme used to compare our model in a benchmark study against other alternative modeling approaches we describe in Section~\ref{sec:modcomp}. In Section~\ref{sec:results} we report the results of these model comparisons. We finally investigate various aspects of our model to further demonstrate how our approach allows for an intuitive interpretation and quantification of uncertainty. 

\subsection{Evaluation scheme} \label{sec:evalscheme}

\begin{figure}[!t]
	\centering
	 \includegraphics[page=1,
	width = \textwidth]{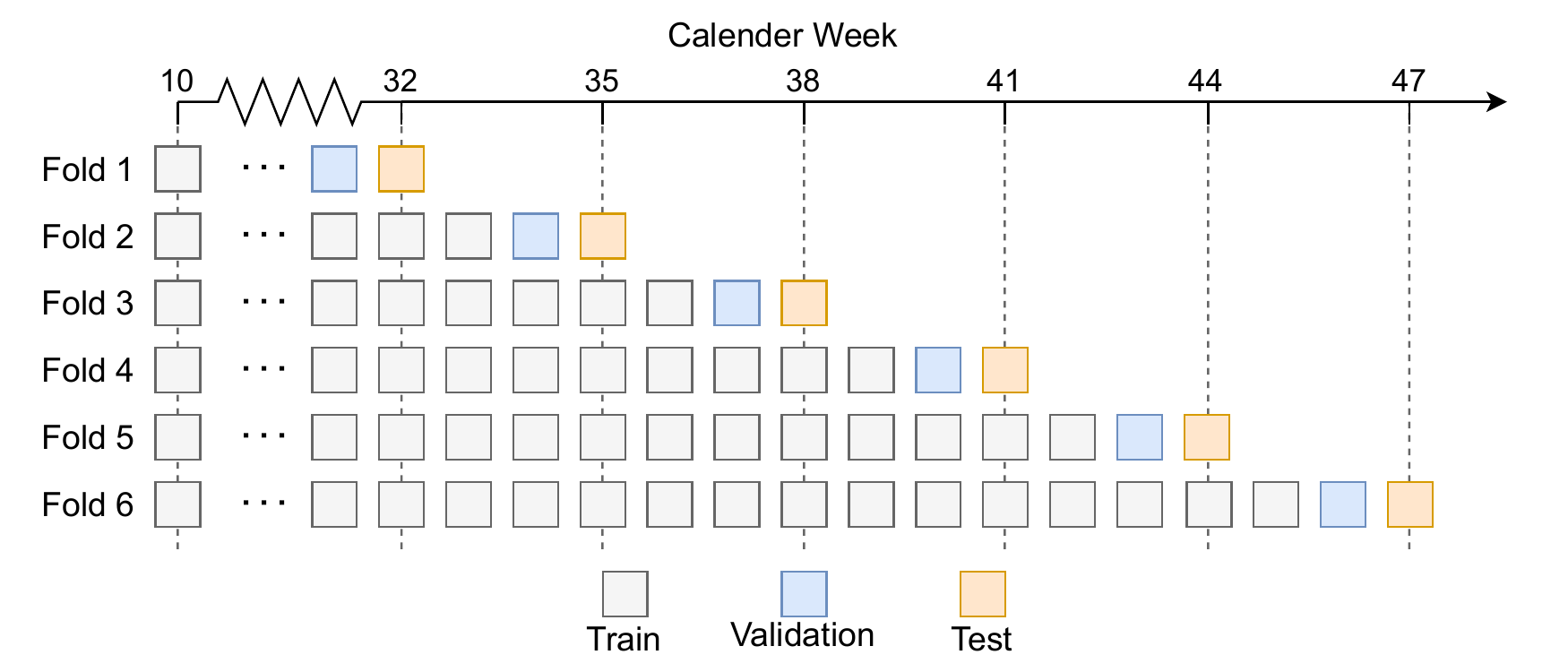}
	\caption{Evaluation Scheme based on an expanding window approach over the available historical weekly data.}
	\label{fig:cv}
\end{figure}

Apart from our primary goal to provide one-week forecasts, we also investigate our approach's behavior throughout the pandemic. Therefore, we apply an expanding window approach, where we use a certain amount of data from past weeks, validate the model on the current week, and forecast the upcoming week. We do this for different time points, starting with training until week 30, validation on week 31, and testing on week 32. In 3-week-steps, we expand the time window while adapting the validation and test week. In Figure~\ref{fig:cv} we sketch a visual representation of this scheme.

\subsection{Model comparisons} \label{sec:modcomp}

We compare our approach against four other algorithms and different model specifications of our framework. As a baseline model, we use the mean of a sliding window approach applied to the given training data set (MEAN). The prediction on the test set then corresponds to the mean of the last week in the training data for each of the four subgroups (age, gender). Our statistical regression baseline is a generalized additive model (GAM) inspired by the work from \cite{Fritz2020}, modeling the mean of a negative binomial distribution using various smooth predictors and tensor-product splines. We further apply gradient boosting trees as a state-of-the-art machine learning baseline. 
Due to its computational efficiency, scalability and predictive performance, we choose the well-known XGBoost implementation \cite{Chen.2016} in our comparisons. Finally, we compare our model against a vanilla deep neural network (DNN), a multi-layer perceptron with a maximum amount of 4 hidden layers with ReLU or tanh activation function, dropout layers in between and a final single output unit with linear activation. To enable a meaningful comparison, we corrected all benchmark model outputs for the differing exposures by incorporating an offset in same way as explained in Section~\ref{sec:nn_form}. Similar to classical statistical models, this allows the model to learn the actual rate of infections. In all cases, we optimize the model using the Poisson log-likelihood (count loss). We furthermore tune the DNN and XGBoost model using Bayesian optimization (BO; \cite{Snoek.2012}) with 300 initial sampled values for the set of tuning parameters and ten further epochs, each with 30 sampled values. Finally, we investigate the performance of a simple GNN, i.e., not in combination with distributional additive regression, optimized on the RMSE. 

\subsection{Results} \label{sec:results}

First, we report forecasting performances on all data folds (based on the optimal setting found by BO). Consecutively, we scrutinize our model's behavior by examining partial effect plots and estimated uncertainties.

\paragraph{Forecasting performance}

In Table~\ref{tab:forecasts} we give the forecast performances of our model and approaches described in Section~\ref{sec:modcomp}. Out of the benchmark models, the GAM is the best performing models returning consistently smaller RMSE values than XGBoost and the DNN, with one exception in week 41. The rolling mean (MEAN) performs similar well before the second wave in Germany (Week 32 and 35). However, the numbers stagnate during Germany's second lockdown (Week 47), which may be seen as an external shock that cannot be accounted for by the previous weeks' rising numbers. The vanilla DNN yields the worst performance, where the BO found the smallest architecture with only one hidden layer with one unit to be the best option. While this result aligns with the good performance of MEAN and GAM, dropout in the DNN between the input and hidden layers does apparently not yield enough or the appropriate regularization to prevent the DNN from overfitting. Our model shows similar performance to the GAM model, which is again in line with our expectations, as we orthogonalize the GNN part of our network w.r.t. the whole structured additive part. 
In particular, the model performs notably better for the weeks 41 - 47 than the GAM, and yields the best or second best results compared to all other models on each fold. Although XGBoost outperforms our approach on week 41, its worse performance on all other folds does not make it a reasonable choice. The same holds for the NB variation of our approach, which delivers the second-best performances. Finally, the GNN itself yields reasonable predictions in the first two weeks but does not perform well for the other weeks. 

\begin{table}[!t]
\begin{small}
\begin{center}
\begin{tabular}{lcccccc}
& \multicolumn{6}{c}{Calendar Week} \\
       & 32 & 35 & 38  & 41  & 44  & 47 \\ \hline
\multicolumn{1}{l|}{XGBoost}      & 4.926 & 5.188 & 7.447 & \textbf{15.327} & 65.036 & 74.235 \\
\multicolumn{1}{l|}{DNN}          & 10.179 & 12.178 & 17.897 & 64.065 & 108.474 & 80.901 \\
\multicolumn{1}{l|}{GAM}          & {4.042} & 4.738 & {4.736} & 21.666 & {18.556} & 23.813 \\
\multicolumn{1}{l|}{MEAN}      & 5.038 & {3.666} & 6.196 & 30.910 & 20.090 & {23.159} \\
\multicolumn{1}{l|}{GNN} & 5.972 & 6.785 & 11.355 & 49.064 & 77.162 & 53.489 \\
\hline
\multicolumn{1}{l|}{Ours (ZIP)} &
\textbf{3.931} & 4.235 & \textbf{4.500} & {16.588} & \textbf{17.738} & \textbf{15.050} \\
\multicolumn{1}{l|}{Ours (NB)} & 4.096 & \textbf{4.094} & 5.174 & 28.580 & 18.098 & 31.724 \\
\end{tabular}

\caption{RMSE values for different methods (rows) and folds (columns) corresponding to the data splitting approach explained in Section~\ref{sec:evalscheme} and Figure~\ref{fig:cv}. Bold numbers denote the best result in each fold across all models.}
\label{tab:forecasts}
\end{center}
\end{small}

\end{table}

\paragraph{Model interpretation}

We now provide further insights into our model by analyzing the partial effects of its structured additive part. More specifically, we focus on two instances of our models that are trained with data until calendar week 30 and 44. We pick the respective weeks to showcase the partial effects during a high and low season of the pandemic. 

\begin{figure}
	\centering
	 \includegraphics[trim={1.6cm 0cm 4cm 2cm},clip,width = 0.49\textwidth]{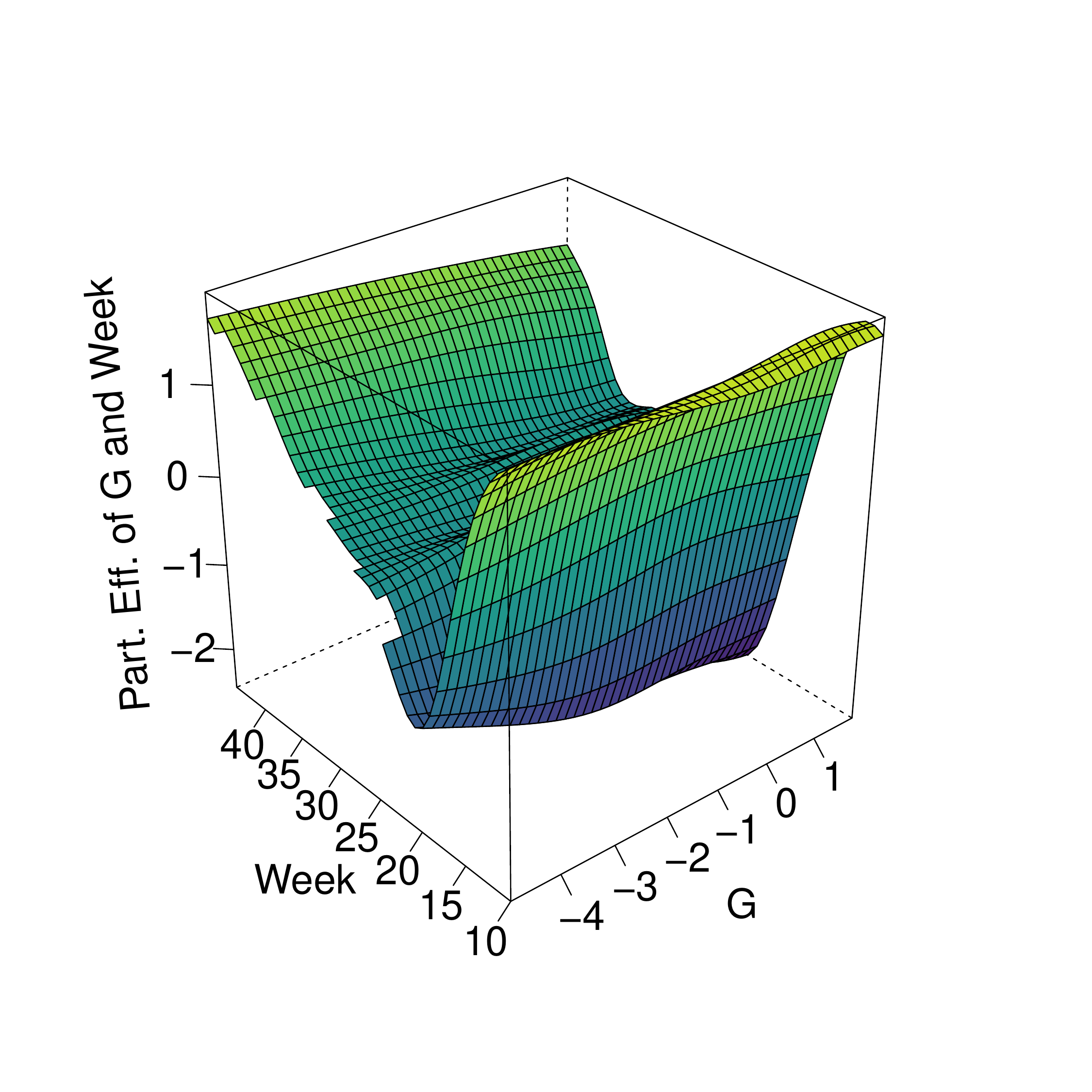}
	 \includegraphics[trim={1.6cm 0cm 4cm 2cm},clip,width = 0.49\textwidth]{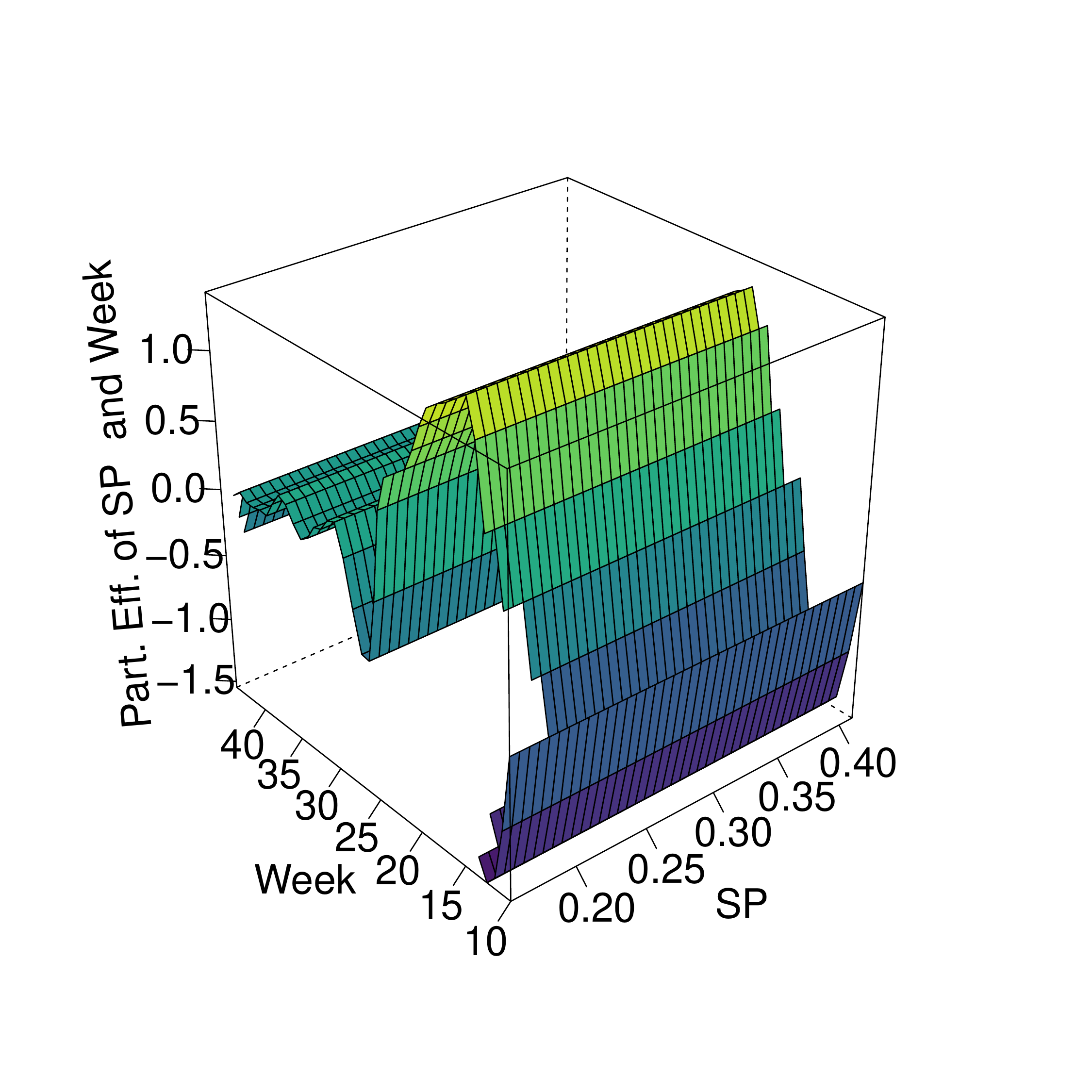}
	 \caption{Estimated bivariate partial effects of the week and Gini coefficient (G) on the left as well as of the week and Percentage Staying Put (SP) on the right.}
	 	\label{fig:bivariate}
\end{figure}

To begin, we investigate the partial effects of the Gini coefficient (G) derived from the colocation maps and the percentage of people staying put (SP) on the left and right side of Figure~\ref{fig:bivariate}, respectively. Moreover, a high standardized Gini coefficient translates to meeting behavior that is more dispersed than the average of all districts. Hence, a low standardized Gini coefficient (less mobility than average) leads to lower infection rates, especially between calendar week 10 and 30. For the percentage of people staying put, the temporal dynamics are somewhat opposite and exhibit small effects in the first weeks and after week 30. Thereby, we may conclude that having a higher percentage of people staying put also lowers the infection rates. Further, we observe that the incorporated penalty term successfully regularizes the bivariate effect term to be a linear effect in the direction of the percentage of people staying put. 

\paragraph{Epistemic uncertainty}

\begin{figure}
	\centering
	 \includegraphics[clip,width = 0.49\textwidth]{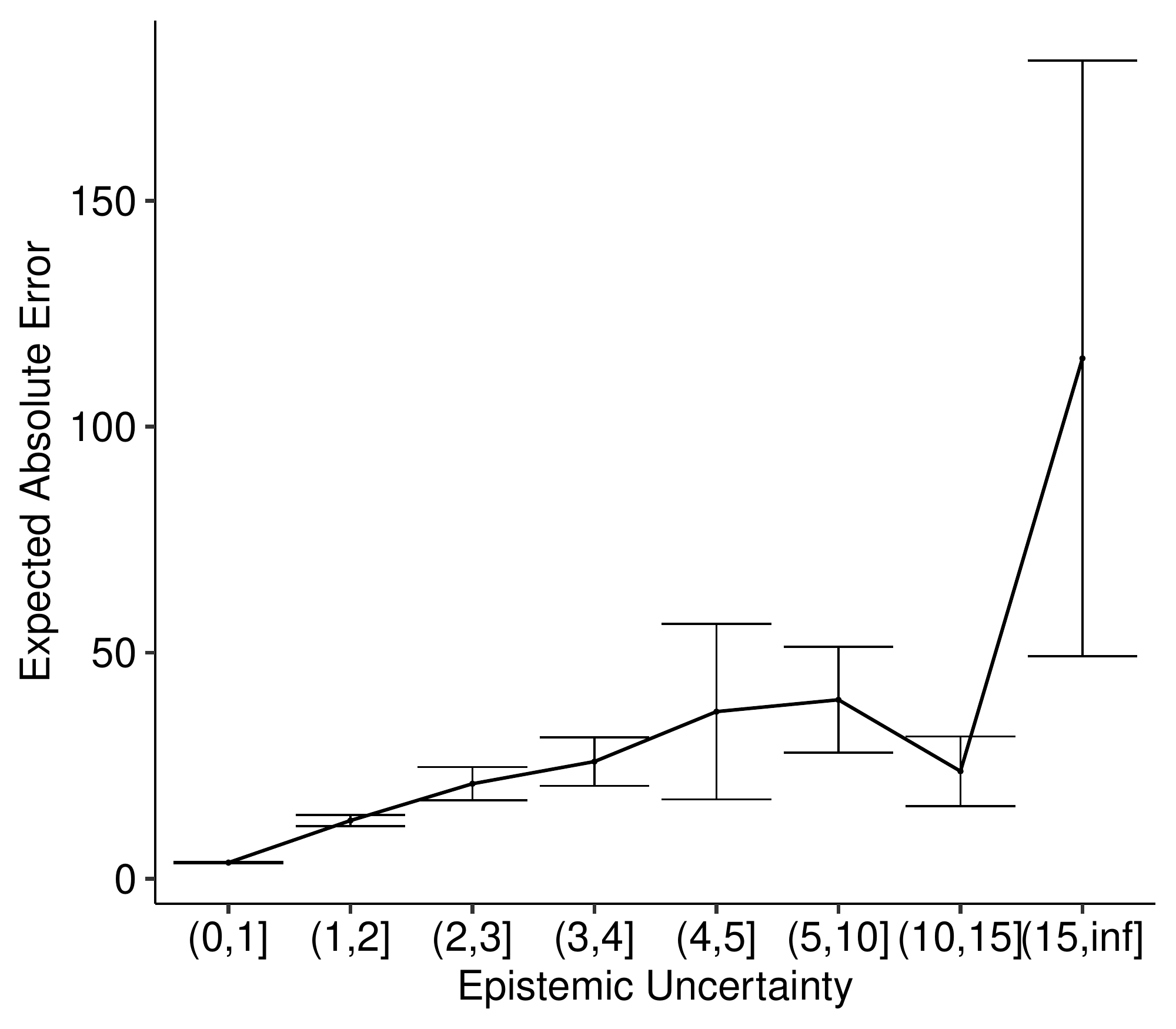}
	 \includegraphics[clip,width = 0.49\textwidth]{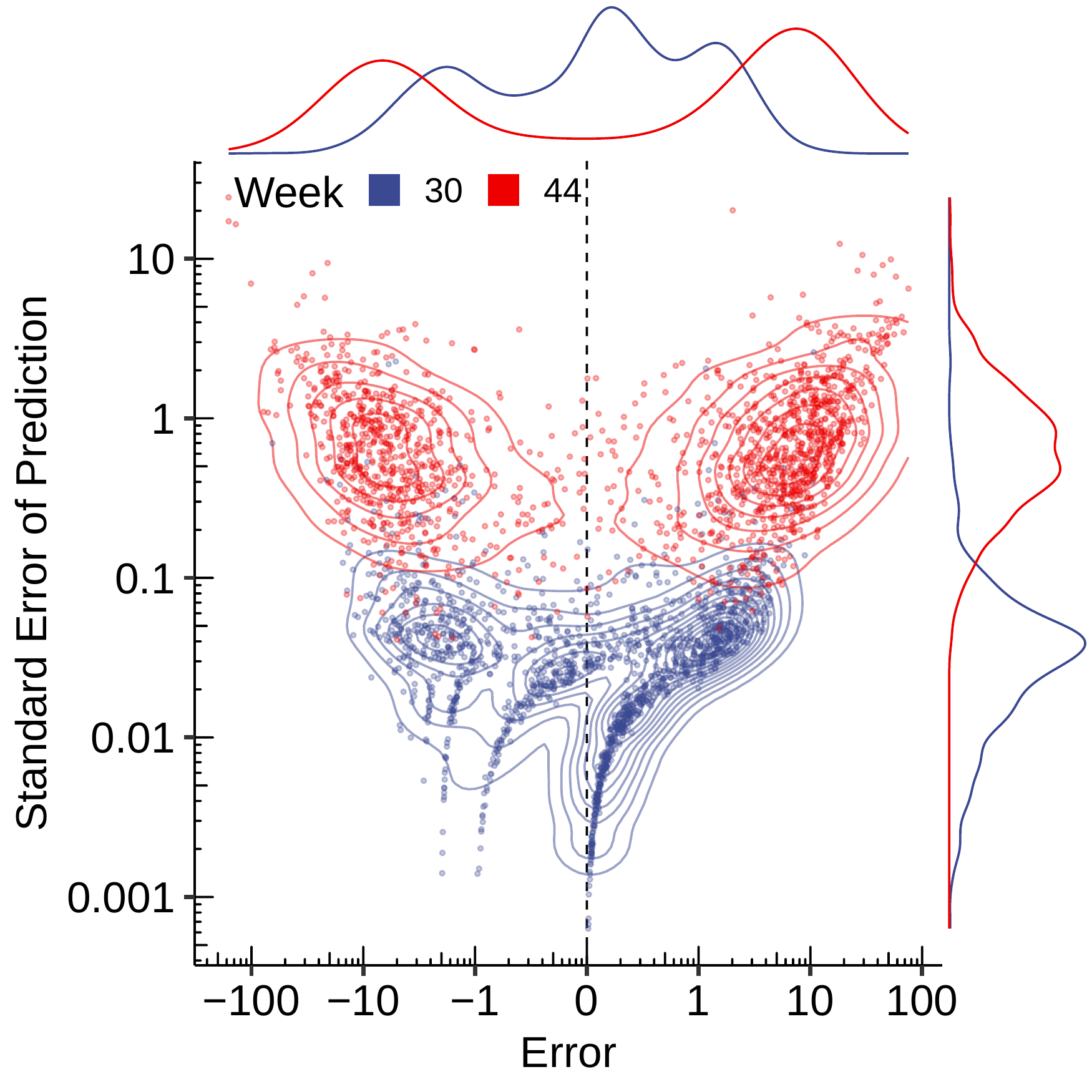}
\caption{
Left: average absolute error incurred by the ensemble for increasing levels of epistemic uncertainty, with vertical bars denoting $\pm$ one standard error of the estimation.
Right: Standard error of the predictions of an ensemble of ten networks correlated with the error incurred when using the ensemble's mean prediction for each district, age, and gender cohort during a low-infection phase (week 30) and the second wave of infections (week 44).
}
\label{fig:epi2}
\end{figure}

As explained in Section~\ref{sec:uq}, while an epistemic uncertainty for the structured part of our model can be derived theoretically, we estimate our models' uncertainty through an ensemble for the GNN part.
Specifically, we repeat the training procedure ten times, each time with a different random initialization. Consecutively, we compute the standard errors of all predictions, thus defining the epistemic uncertainty.

The epistemic uncertainty is well correlated (Spearman's $\rho=0.76$) with the absolute error resulting from the mean prediction (Figure~\ref{fig:epi2} left) and grows approximately linearly with the error.
However, the variability of the average error is not reliable in the last bin due to the low number of samples with high ($>15$) epistemic uncertainty.
The epistemic uncertainty is generally higher for high-incidence weeks such as week 44 compared to a low-incidence week such as week 30 (Figure~\ref{fig:epi2} right).
In addition, the ensemble has a very slight tendency to underestimate the number of cases for week 44 by 1.26, and to overestimate the cases for week 30 by 0.25.
Although statistically significant (one-sided $t$-test, $t=3.24$, $p=0.001$ and $t=3.38$, $p=0.0007$, respectively), the resulting differences are practically irrelevant, hence suggesting that the ensemble is approximately calibrated.
In general, this correlation between epistemic uncertainty and forecast error provides a reliable diagnostic of the trustworthiness of our proposed model's predictions.

\begin{figure}
	\centering
	\includegraphics[width = 0.5\textwidth]{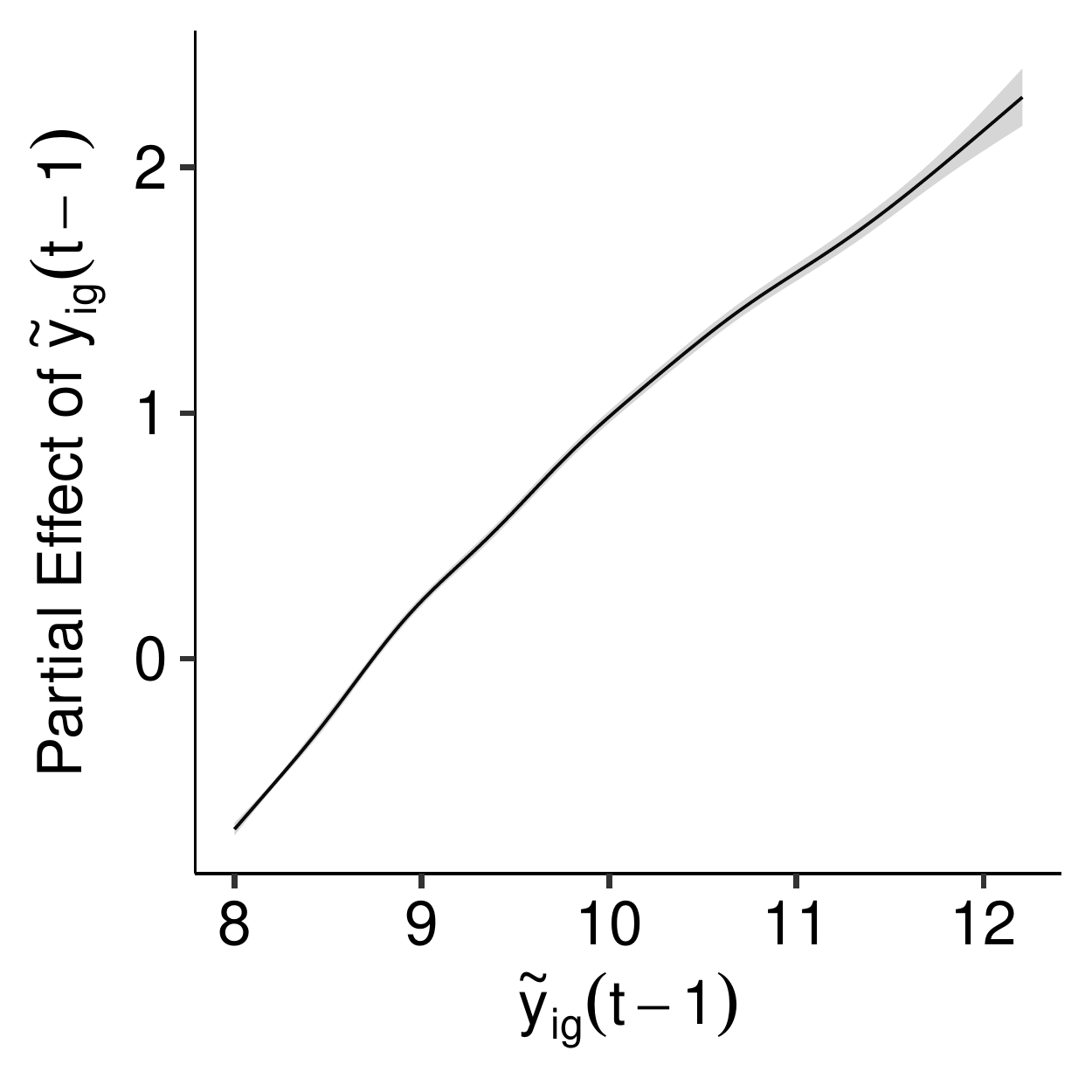}
	\caption{Estimated effect of lagged infection rate with corresponding confidence intervals for week week 44.}
	\label{fig:epi1}
\end{figure}

The partial effect of lagged infection rates in Figure~\ref{fig:epi1} additionally depicts its epistemic uncertainty when the GNN weights are fixed. The figure's narrow shaded areas translate to high certainty of the partial effect from the respective feature. Moreover, the partial effect translates to the finding that the higher the infection rate was in the previous week, the higher the predictions are in the upcoming week. In line with this result prior studies, already identified this feature as a principal driver of the structured model part of our network \cite{Fritz2020}. 

\paragraph{Aleatoric uncertainty}

\begin{figure}
    \centering
    \includegraphics[width=\textwidth]{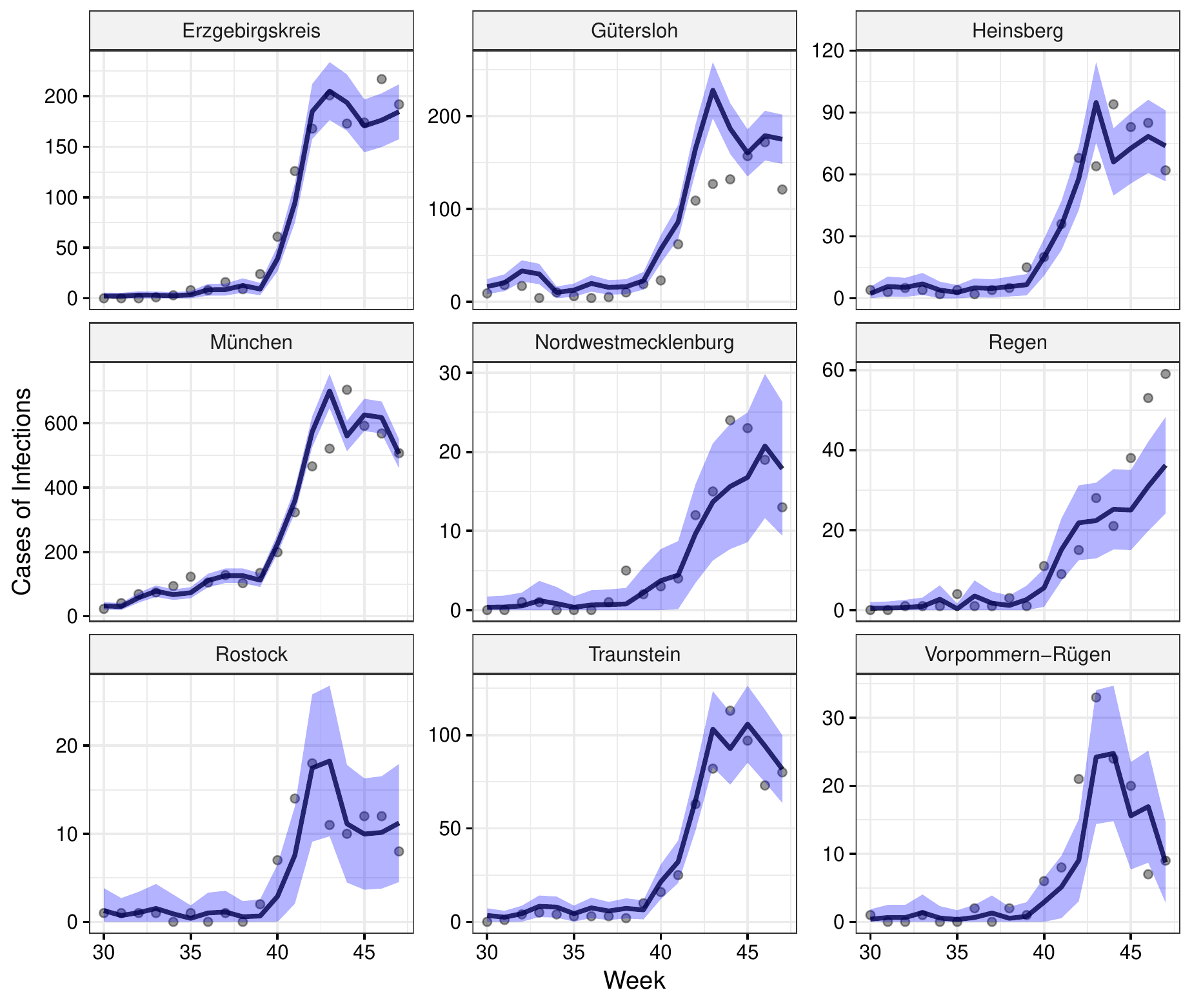}
    \caption{Exemplary prediction intervals for age group 35 - 59 and gender male and selected districts (facets) in Germany. For the different forecast weeks (x-axis) the true number of infections are visualized by points and contrasted with the model's prediction (black line) and the prediction interval (shaded purple area).}
    \label{fig:alea2}
\end{figure}

We evaluate our ZIP model's aleatoric uncertainty by applying the expanding window training scheme analogous to our model evaluation. For each prediction, we calculate predictive distribution intervals using the mean prediction $\pm$ 2 times the standard deviation derived from \eqref{eq:aleatoric}.
Figure~\ref{fig:alea2} depicts the probabilistic forecasts of the modeled ZIP distribution for different districts in Germany. These districts constitute particularly difficult examples from relatively rural areas and cases in larger cities (Munich / M\"unchen) as well as sites that were hardly affected and severely affected by the pandemic. In Figure~\ref{fig:alea2} we visualize the true values as points and the predicted mean as a black line. Here, the shaded purple area symbolizes the predictive distribution intervals. We observe that most of the points are well within the given prediction interval, thus the distribution captures the dispersion in the data quite well. As expected from the Poisson distribution, results indicate that the aleatoric uncertainty increases with the rising number of infections. However, some intervals are not able to cover larger fluctuations and steeper increases of infections, such as is the case in \emph{M\"unchen}, \emph{G\"utersloh} or \emph{Vorpommern-R\"ugen}. 

Overall, the intervals derived from the predictive distributions cover on average over 80\% of all cases in all groups, weeks and districts. This indicates that the estimation of distribution variances for groups and weeks works well, but shows also room for improvement for later weeks where the distribution is not perfectly calibrated. 


\section{Discussion}
\label{sec:discussion}
Following several experts' call to account for human mobility in existing statistical and epidemiological models of COVID-19, we propose a multimodal network that fuses existing efforts with graph neural networks. We thereby enable the use of a more nuanced collection of data types, including mobility flows and colocation probabilities, in the forecasting setting. Our results indicate a notable improvement over existing approaches, which we achieved by accounting for the network data. The given findings also highlight the need for regularization and showcase how common ML approaches can not adequately capture the autoregressive term, which, in turn, proved to be essential for the forecast. Our model's investigation further showed that uncertainty can be well captured by the model, although further calibration may be vital for its aleatoric uncertainty.

\paragraph{Caveat} We want to emphasize that despite the convincing results, the given analysis only addresses a small subset of processes involved in the spread of COVID-19 and should not be the sole substance for decision making processes in the future. In particular, forecasting infection rates in the short run does not (need to) address reporting or observation biases typically present for such data and requires a solid data basis, which in the case of Germany, is provided by the Robert-Koch Institute.


\section*{Acknowledgements}
The work has been partially supported by the German Federal Ministry of Education and Research (BMBF) under Grant No. 01IS18036A. ED is supported by the Helmholtz Association under the joint research school "Munich School for Data Science - MUDS" (Award Number HIDSS-0006). The authors of this work take full responsibilities for its content.

\bibliography{mybibfile}

\end{document}